# Winning Solution of the AIcrowd SBB Flatland Challenge 2019-2020

Mugurel-Ionut Andreica

## 1) Overview

This report describes the main ideas of the solution which won the [AIcrowd SBB Flatland Challenge 2019-2020](), with a score of 99% (meaning that, on average, 99% of the agents were routed to their destinations within the allotted time steps). The details of the task can be found on the competition's [website]().

The solution consists of 2 major components:
1) A component which (re-)generates paths over a time-expanded graph for each agent
2) A component which updates the agent paths after a malfunction occurs, in order to try to preserve the same agent ordering of entering each cell as before the malfunction. The goal of this component is twofold:
    a) to (try to) avoid deadlocks
    b) to bring the system back to a consistent state (where each agent has a feasible path over the time-expanded graph)

I am discussing both of these components, as well as a series of potentially promising, but unexplored ideas, below.

## 2) (Re-)generating agent paths

The invariant for this component is that every agent always has an assigned path (where it will be located at each time step over the whole time horizon), and this component only tries to improve the overall path assignment).

Initially, all the agents have a default path assigned which doesn't enter the environment at all (they always just stay at their initial location, outside the environment).

A series of up to **max_runs** are run. At each run **num_threads** threads are running in parallel, each of them generating **num_permutations** random permutations of agents (see more details below). For each permutation of agents a path (re-)assignment is computed and the best path assignment over all of the **num_threads x num_permutations** such assignments is kept. If the initial path assignment is improved during a run, then the best found path assignment is saved and used as the initial assignment for the next run. Otherwise the next run is not even started.

An agent permutation is generated with the following conditions: in decreasing order of their speed **S**, all agents with the same speed **S** are randomly shuffled and added to the current permutation (thus, all agents with speed 1 are added before all agents with speed 0.5, etc.).

An agent permutation **a(0), …, a(N-1)** is handled as follows. Consider the agents in the order given by the permutation. First, the current agent path is removed from the time-expanded configuration and then a shortest path algorithm is run in order to find a new path which doesn't interfere with the current paths of all the other N-1 agents.

The time-expanded graph consists of all possible tuples **(cell, orientation, time)**. Orientation is between 0 and 3, time is up to ~2800 (larger than the maximum possible real time horizon) and cell is an index up to the maximum number of cells occupied by a railroad piece. Although the maximum dimensions of a flatland board are 150x150, implying that there can be up to 22500 cells occupied by railroads, in reality most of them are empty. Based on local tests with various arguments of the sparse rail generator, up to 3000 cells are occupied in any test. This was also the case for all of the 250 official tests.

A data structure marking for each **(cell, time)** tuple which agent occupies it during time step **time** (or if it's free) is also maintained. In fact, removing a path from the configuration means marking all the **(cell, time)** tuples of the path as free. This data structure is used by the algorithm which tries to find a new shortest path to its destination for the current agent A of the permutation.

The implemented shortest-path algorithm is essentially an A*-style type of algorithm, where tuples are added to a heap and sorted according to the estimated time to reach the destination (this can be computed correctly because shortest paths from each destination to each **(cell, orientation)** are pre-computed). In case of ties, tuples corresponding to paths in which the agent enters the environment later are preferred.

Only **(cell, orientation, time)** tuples from which the agent can actually make a decision going forward are added to the heap. For instance, if the agent needs **T** turns to move to an adjacent cell and starts a move from the cell **C** at time **t1**, then the intermediate tuples **(cell, t1+x) (1<=x<=T-1)** are not added to the heap.

The data structure which marks occupancies (together with the rules regarding the order in which agent moves are performed at each turn and the fact that agents exit the environment as soon as they enter their destination cells) is used to decide if an agent can move to an adjacent cell or if it can stay one more turn at the current cell.

If an agent is already inside the environment and is executing an ongoing move at the current time step of the simulation, then the move is simulated until the agent enters its target cell (or, if the target cell is blocked, until the last time step of the time horizon is reached). And only afterwards will the heap-based algorithm start. This is because agents cannot be stopped once they start a move, until they reach the adjacent cell towards which they started moving. Also, if the agent is currently malfunctioning, then this is also taken into account (meaning the heap-based shortest path algorithm is only run from a configuration where the agent is not moving and is not malfunctioning anymore).

Please note that the shortest-path algorithm is always optimistic, meaning it doesn't consider any future malfunctions when planning the agent paths (only the ongoing malfunctions are considered).

The score of a path assignment is a tuple **(X, Y)**, where **X** = the number of agents who successfully reached their destination (this number is usually 100% because I used a time horizon larger than the real one), and **Y** is the sum over all agents **A** that reached their destination over **TD(A)$^E$**, where **TD(A)** is the time step at which agent **A** reaches its destination and **E** is an exponent. I experimented with multiple values of **E** ranging from **0.25** to **4.0**, but the best results were obtained for **E=1** (which corresponds almost exactly to maximizing the sum of rewards provided by the Flatland environment).

There is also an additional piece of logic which tries to reduce the chance of future deadlocks (or completely avoid them), but I will discuss this part in the following section.

## 3) Updating agent paths after a malfunction

Whenever an agent starts malfunctioning, the path assignments that have been computed previously can become invalid. This causes problems from two points of view:
1) The paths cannot be used as they are for deciding the moves each agent makes at each step
2) The paths cannot be used as they are as initial path assignments for the component which re-generates paths, because that component requires an initial consistent (i.e. valid) path assignment

So it is of paramount importance to modify the current path assignment into a consistent (valid) path assignment (even one with a lower quality), because only that can then be used as the starting point for the path re-generation logic.

## 3.1) The core algorithm

Here I used existing ideas from the literature as a starting point. In [1] and [2] the concept of **plan-step graph** is defined. Essentially, for every cell **C** of the environment, an ordering **a(C, 1), a(C, 2), …, a(C, num_cell_visits(C))** is defined, corresponding to the order in which agents enter the cell (note that it's allowed for the same agent to enter the cell multiple times, if needed). Each visit of an agent, **a(C, i)**, has an associated time **tc(C, i)**=the time step when the agent enters the cell.

The same ordering can be viewed from the agents' perspective. Each agent A has an ordering **c(A, 1), c(A, 2), …, c(A, num_agent_visits(A))**, where c(A,i) is the cell visited by the agent A during its i-th move. These visits also have corresponding times: **ta(A,i)**=the time step when the agent A enters the cell **c(A,i)**. Additionally, we can maintain the direct correspondence between agent and cell visits: **agent_to_cell_visit(A, i) = j**, where the i-th visit of agent **A** (to cell **c(A, i)**) is the j-th visit at that cell (i.e. **a(c(A,i), j) = A** and **tc(c(A,i), j) = ta(A,i)**). Similarly, a **cell_to_agent_visit(C, j) = i** mapping can be computed.

When an agent malfunctions we can try to also delay all the visits that are impacted by the agent's malfunction, such that the ordering of the visits in each cell is preserved. The drawback, of course, is that the lengths of the paths are increased, but this is not necessarily an issue, because we can re-route some of these paths using the path re-generation algorithm.

The biggest problem encountered at this point is caused by agents which are already moving when a malfunction of some other agent occurs. If we could introduce delays (i.e. wait moves) at any time step to any agent, then preserving the visiting order is trivial and can be performed by a simple traversal (BFS or DFS) of the precedence graph defined by the ordering of the visits.

The precedence graph has a node for each visit. Then, there is a direct arc from a visit **(C1, i1)** to a visit **(C2,i2)** if:
- **C1** is the destination cell of the agent **a(C1,i1)** and **i2=i1+1**
- **a(C1,i1) = a(C2,i2-1)** and **cell_to_agent_visit(C2,i2-1) + 1 = cell_to_agent_visit(C1, i1)**
    - Briefly, this means that **(C1, i1)** is the next visit of the agent corresponding to the visit **(C2, i2-1)**. This follow-up visit must start (in order to free the cell **C2**) before the visit **(C2, i2)** can start.

If an agent is already moving, then this agent cannot be paused until it reaches the adjacent cell towards which it is moving. This means that if two agents move towards the same cell **C**, and are scheduled to visit it in the order **A1**, followed by **A2**, then if **A1** malfunctions before it enters **C** but after **A2** started moving towards **C**, then the ordering of the visits cannot be preserved. Not being able to preserve the ordering of the visits can introduce deadlocks. Once an agent is part of a deadlock, then this agent might be blocked until the end of the simulation. I am only saying "might", because it's possible that the agent can be unblocked by re-routing it on a different path - however, there's no guarantee of such a possibility.

Moreover, having a deadlock blocks not also agents, but also the cells occupied by deadlocked agents, potentially causing major issues to the non-deadlocked agents.

Let's ignore the potential for deadlocks at the moment. My implementation of the algorithm for updating agent paths used BFS. A queue is used, containing **(tmin, A, C, aidx, cidx)** tuples, where:
- **tmin** is the minimum time at which agent may enter cell
- **A** is an agent id
- **C** is a cell id
- **aidx** is the index of the next move of the agent (it should be a move to the cell **C**)
- **cidx** is the index of the next visit taking place in the cell **C**; it should correspond to the visit by the agent **A**

Initially the tuples corresponding to the next move of each agent (if any) are added to the queue.

Based on the time when the agent entered its current cell, the agent's speed, whether the agent is malfunctioning and/or is in an ongoing move, and based on the value of **tmin**, the earliest time **T** when the agent can enter the cell **C** is computed. If **T** is less than **tmin**, then an inversion of the visiting order occurred (this can only happen if the agent is part of an ongoing move and it can enter the cell **C** before **tmin**).

If the move can be performed successfully, then two new tuples may be added to the queue:
1) the one corresponding to the next move of the same agent (i.e. the visit with index **aidx+1** of agent **A**, if such a visit exists)
2) the one corresponding to the next agent who wants to enter the agent's current cell (i.e. the cell from which the agent **A** starts moving in order to reach **C**)

For the 2nd tuple, **tmin** is computed such that the other agent only enters the agent's current cell after **A** left it (also considering the rules regarding the order of performing agent moves during each simulation step).

If the move can only be performed before **tmin**, then a very basic attempt of changing the ordering of the two involved agents is performed (i.e. to swap them in the ordering of the conflicting node and then run the whole algorithm again), but this doesn't guarantee a deadlock won't occur. If a deadlock occurs, then this will be noticeable because not all plan-steps will have been performed by the algorithm.

In this case, one option is to just mark the deadlocked agents as deadlocked and simply continue with the remaining agents. The path assigned to a deadlocked agent will consist of staying in the same cell until the end of the simulation (or to move - unsuccessfully - towards the cell towards which it's already moving). This path can be modified later by using the path re-generation component.

An early implementation of my solution did just that, and was able to bring 96.5%[*] of the agents to their destinations, on average, on the official 250 tests. This was my first submit that didn't fail (all the previous failures were caused by various bugs and adding many asserts in the code to check that the simulation environment is behaving exactly as expected by the planning logic).

## 3.2) Avoiding deadlocks

The next logical approach was to try to fully avoid deadlocks. As mentioned, a deadlock can only occur if there are 2 agents **A** and **B**, both moving towards the same cell **C**, and the time intervals of their (ongoing) moves overlap. A time interval of an (ongoing) move is an interval **[t1, t2)**, meaning that the agent started moving at time **t1** from its current cell, and reached the target cell at time **t2**. At best, **t2=t1+1** (for agent with unit speeds which are not impacted by a malfunction during the interval). However, because of different speeds and/or ongoing malfunctions, it's quite possible that **t2>t1+1**.

So, if the move intervals of 2 agents **A** and **B** moving towards the same cell **C** overlap, and the agent which is supposed to arrive first at **C** malfunctions during the time window of overlap, then the other agent cannot pause and will simply reach **C** before the malfunctioning agent (causing an inversion in the visiting order, which in turn may lead to deadlocks).

The simplest solution is to avoid having 2 agents moving towards the same cell during overlapping time windows. This needs to be enforced both during the path re-generation phase, as well as during the phase which updates paths after a malfunction (by assigning **tmin** values to tuples which ensure that consecutive visits to the same have fully non-overlapping move time intervals).

Implementing this approach indeed avoided deadlocks in all the cases and I could tune it to get an average of ~97.5%[*] of agents reaching their destinations.

Unfortunately, as things stood, there wasn't a lot of room for improving the solution further. The problem was that now the agents were spaced too far apart in time, reducing the available "throughput". There are 2 inefficiencies here:
1) Not all pairs of agents **(A,B)** with overlapping time intervals moving towards the same cell **C** will cause a deadlock even if the order of their visits to **C** is reversed. In order for a deadlock to occur, the plan-step precedence graph must contain a cycle after reversing their visiting orders. This is not necessarily always the case.
2) Even if the potential for deadlock exists if the visiting order of a pair of agents **(A,B)** moving towards the same cell **C** is reversed, an actual deadlock will only occur if the agent arriving earlier at **C** malfunctions during the time window of overlap. Such a situation, while possible, may be statistically improbable in some cases.

Because 1 seemed hard to implement correctly and efficiently enough, I chose 2. I heuristically relaxed some of the time window overlap constraints and ran many local tests. I chose a combination which never caused any deadlocks during my local tests, although I believe that deadlocks are still theoretically possible (under the right circumstances), though unlikely. So, in principle, these constraints can be considered speculative because they don't provide a full guarantee of deadlock avoidance. In case a deadlock does occur, it can be handled as already mentioned in Section 3.1. However, in the final submission, I actually use asserts to ensure that no deadlock actually occurs.

The specific constraints that I kept are the following:
- Avoid nested time intervals of agents going to the same cell

(*) I had a bug in the heap implementation used by the shortest path algorithm, which I only discovered towards the end of the contest. When I discovered and fixed it, my submissions which were scoring 98.2% at the time immediately jumped to 98.9%. This bug existed in my code from the very beginning (the heap implementation was one of the 1st components I wrote), so the reported scores are also affected by this bug. Thus, it's possible that the scores reported here are also affected by a similar 0.7% difference (so a 96.5% score should in fact be 97.2%, and a 97.5% score should in fact be 98.2%).

## 4) How it all fits together

In the first time step of the simulation the path re-generation logic is run in "initial mode" (**max_runs=4, num_threads=3, num_permutations=20**). So a path is assigned to each agent. Then, whenever a malfunction occurs at some time step (even if there are more agents malfunctioning at the same time), the logic to update agent paths is run.

After running the path update logic, the path-regeneration logic is *sometimes* run. If the path update logic increased the maximum time when an agent reaches their destination (or the number of agents reaching their destination decreases), then a counter is increased. If this counter exceeds **3**, then the path-regeneration logic is run in "full mode" (**max_runs=4, num_threads=3, num_permutations=10**), and then the counter is reset; otherwise, the path re-generation logic is run in "restricted mode" (**max_runs=2, num_threads=3, num_permutations=2**).

Ideally, the path re-generation logic would be run as often as possible (e.g. "full mode" after every path update) with **max_runs**, **num_threads** and **num_permutations** as large as possible. The parameters mentioned above were chosen so that the official 250 test cases can be handled in less than 8 hours.

# 5) Unexplored ideas

Despite this being the winning solution, there are still multiple unexplored ideas which I believe could have improved the solution quality even further, but which would have required a non-trivial amount of time to implement and tune correctly, while at the same time having uncertain outcomes (i.e. it's not sure the quality of the solution would really increase, or that the new approach would fit within the time limit).

1) An idea for maximizing "throughput" (agents traveling to their destinations) while guaranteeing deadlock avoidance, was to construct the plan-step precedence graph and use it during the algorithm for finding a shortest path in the time-expanded graph for a given agent **A**. Basically, when considering a move towards a cell **C** during a time interval **[t, t+S)**, the idea was to look at all the overlapping intervals of other agents also traveling towards **C**, construct the plan-step graph by also adding the visit of **A** to **C** at the corresponding time, and evaluate whether any cycle might be introduced in this graph in case the relative order of agent A and any of the overlapping agents is changed (by permuting these agents in a different order consistent with a potential set of malfunction durations for a subset of these agents). If none of the considered reorderings could cause a cycle in the plan-step precedence graph (or the probability of obtaining a cycle is estimated to be sufficiently low), then the move would be considered safe to use along agent A's new path.

One problem with this idea is that it could potentially slow down the shortest path finding algorithm quite a lot, which was already the bottleneck of my solution. This could have caused my solution to not fit within the 8 hour time limit for the 250 official test cases.

2) Updating the agent paths after a malfunction doesn't necessarily need to preserve the visit ordering. I was thinking of trying a local search type of algorithm over the visit ordering, randomly changing the order of some visits, then run the algorithm to assign times to the visits (and detect deadlocks) for each changed order, and find the ordering which provides the best score. A move (change of ordering of a subset of visits) could be accepted either only if it improves the score (hill climbing), or using a simulated annealing type of acceptance condition.

The problem here is that changing the visiting order of 2 agents at a cell C may not be enough and it may be necessary to also change their order in multiple other cells to avoid deadlocks. Detecting which other order changes are needed is a difficult problem, making an approach based on random moves unlikely to find good or even feasible solutions.

3) Based on past malfunction data, estimate the chance of a deadlock, or the expected number of agents to be involved in a deadlock, or even just the chance of having any visit ordering reversals, in case a certain move is chosen for an agent **A** (e.g. moving towards cell **C** starting at time **T**) during the shortest path finding algorithm. Then allow the move only if this number is low enough.

4) Instead of re-generating agent paths one at a time in some order (as the path re-generation component does), choose multiple small subsets of agents and generate the paths for all agents in a subset together (as a path in a multi-dimensional time-expanded graph, with each dimension corresponding to an agent in the subset).

A similar idea (to group agents together) was proposed in [3] (though the grouping logic and shortest path algorithm are completely different). As drawbacks, getting the subsets right and pruning the shortest-path algorithm enough to fit within the time limit could have proved very challenging.

5) Using past malfunction data try to estimate the parameters of the malfunction generation logic (e.g. the parameters of the probability distribution), and use them to simulate **F** future scenarios. Then run the path re-generation logic for each of the **F** future scenarios, and pick for each agent a set of paths which maximizes the expected score over all the future scenarios.

As drawback, this is hard to formalize. When there are different paths chosen for each agent depending on the simulated future outcome, it's not clear how to combine them into an "average move" which maximizes the expected score.

One idea would be to just optimize some path re-generations params (e.g. spacing between agents, how much overlap of move time windows is allowed, etc.), and then use these params during the real simulation.

Along the same idea, some default values for these params can be learnt by running many simulations offline and choosing the combinations of params which leads to the best average offline score. Such an approach would essentially combine reinforcement learning (learning hyper-parameters) with operations research / global optimization approaches (where the learnt parameters would be used).

6) Compress the time-expanded graph. Instead of having a node for each (cell, time) pair, we can have a node for each (cell, maximal free time interval) pair. This idea has been described in [4].

# 6) References

[1] ter Mors, A., Witteveen, C., "Plan Repair in Conflict-Free Routing", 2009. In: Chien BC., Hong TP., Chen SM., Ali M. (eds) Next-Generation Applied Intelligence. IEA/AIE 2009. Lecture Notes in Computer Science, vol 5579. Springer, Berlin, Heidelberg

[2] Maza, S., Castagna, P., "A performance-based structural policy for conflict-free routing of bi-directional automated guided vehicles", Computers in Industry 56(7), pp. 719–733, 2005.

[3] Stanley, T., "Finding Optimal Solutions to Cooperative Pathfinding Problems", Proceedings of the 24th AAAI Conference on Artificial Intelligence (AAAI), pp. 173–178, 2010.

[4] ter Mors, A., Zutt, J., Witteveen, C., "Context-Aware Logistic Routing and Scheduling", Proceedings of the 17th International Conference on Automated Planning and Scheduling (ICAPS), 2007.